\documentclass[11pt]{article}

\usepackage[letterpaper,portrait, margin=1.5in]{geometry}
\usepackage{ifpdf}
\ifpdf 
    \usepackage[pdftex]{graphicx}   
    \usepackage{tikz-qtree}
    \usepackage{url}
    \usepackage[pdftex,     
            plainpages=false,   
            breaklinks=true,    
            colorlinks=true,
            pdftitle=My Document
            pdfauthor=My Good Self
           ]{hyperref} 
    \usepackage{thumbpdf}
\else 
    \usepackage{graphicx}       
    \usepackage{hyperref}       
\fi 

\title{PETRARCH 2 : PETRARCHer}
\author{Clayton Norris}
\date{}

\begin{document}
\maketitle
\href{https://github.com/openeventdata/petrarch2}{PETRARCH 2} is the fourth generation of a series of Event-Data coders stemming from research by 
Phillip Schrodt\footnote{Schrodt, 2001}. Each iteration has brought new functionality and usability \footnote{Schrodt, 2012}, and this is no exception.
Petrarch 2 takes much of the power of the original Petrarch's dictionaries and redirects it 
into a faster and smarter core logic. Earlier iterations handled sentences 
largely as a list of words, incorporating some syntactic information here and 
there. Petrarch 2 (henceforth referred to as Petrarch) now views the sentence entirely on the syntactic level. It 
receives the syntactic parse of a sentence from the Stanford CoreNLP 
software, and stores this data as a tree structure of linked nodes, 
where each node is a Phrase object. Prepositional, noun, and verb phrases each 
have their own version of this Phrase class, which deals with the logic 
particular to those kinds of phrases. Since this is an event coder, the core of 
the logic focuses around the verbs: who is acting, who is being acted on, and 
what is happening. The theory behind this new structure and its logic is founded 
in Generative Grammar, Information Theory, and Lambda-Calculus Semantics. 

\section{Tree structure}
In an attempt to take advantage of all the syntactic information provided to us 
by the Stanford Parse, Petrarch implements the sentence coding in such a way that 
the syntax tree is apparent in the data structure and the logic. It's a simple tree 
structure, with every phrase or word being its own node, with pointers to the 
parent node and the set of children nodes\footnote{The relationship between nodes 
is frequently described using terms of familial relation (most frequently, ``parent'' and ``child''), or direction (``above'' meaning ``closer to the 
root,'' and ``below'' meaning ``closer to word-level.'')}. 
The syntactic phrases are stored as nested objects of the ``Phrase" 
class, which has three subclasses ``NounPhrase," ``VerbPhrase," and ``PrepPhrase." 
If a phrase doesn't fall under one of these categories, it is just kept as a 
``Phrase," though if eventually enough reason can be shown to add another type (adjective phrases, for example),
it an be done so easily. Each of these phrase types has several methods unique to it 
and its own version of a $get\_meaning()$ method, which is what determines the 
coding of a node from the meaning of its children. The simplicity of this recursive approach in comparison 
with the expensive list-based pattern matching from previous versions of Petrarch yields a 
significant speed improvement, 
and the theoretically-grounded tree-based semantic searching takes advantage of 
the relationships between nouns and verbs encoded in the syntax tree. 
\subsection{Syntax trees}
Let's start with a short linguistics lesson. Every sentence in English (and most languages) is made up of several ``constituents''. A constituent can be a single 
word or a whole phrase (which is a constituent of constituents), but the defining 
characteristic is that each constituent serves a specific syntactic (i.e. grammatical) role. 
Constituents of a sentence are associated hierarchically (hence the phrasal constituent-of-constituents),
 and so the most convenient way of visualizing or storing syntactic structure 
is in a syntax tree. There is an example of a syntax tree and how it is used in 
the parse at the end of this document. 

The CoreNLP software on which Petrarch relies for syntactic parsing
uses the Penn Treebank II \footnote{\url{https://www.cis.upenn.edu/~treebank/}} syntax notation, which can differ slightly from canonical 
generative grammar labeling, but for our purposes they are equally useful. 
Constituents have specific type, depending on their ``head'' and distribution. The cases we care most about in this program are Noun, Verb, and 
Prepositional phrases. Heads, in this case, are effectively the single word that a phrase 
can be reduced to, both semantically and syntactically. They can be predictably located 
by navigating the syntax tree, so Petrarch relies on the idea of phrasal headedness 
for much of its speed. A head of a phrase can be formalized as the lowest word-level constituent
to which there is an unbroken path of phrase-level similarly-typed constituents from the phrase's root 
node. Basically, to find the head of an NP (Noun Phrase), you follow the path of NP's down the 
tree until you find a Noun. If there's ever a choice of which path to take, in English you will
take the rightmost\footnote{Many theories of syntax dictate that any node can have at most two children, which would 
never yield a situation where you have several choices, but CoreNLP does not follow this binary branching restriction}.

 \section{Flow}
The core logic of the semantic parsing is based on the notion that each node in the 
tree has a meaning, and the meaning of a node is a combination of the meanings 
of its children. That means that in moving up the tree and going from word-level to 
sentence-level, words and meanings get combined until you have one noun phrase 
meaning and one verb phrase meaning. The meaning of the verb phrase is what captures 
most of the meaning of the sentence, and accounts for all the relevant nouns and 
verbs below it. 

Because of the recursive nature of the meaning determination, one call to 
$get\_meaning()$ from the upper most verb phrase will cause a domino effect that 
finds the meaning of the rest of the tree. The flow of each specific phrase type 
is determined by its $get\_meaning()$ method. While the logical flow can't be 
strictly linearized due to the domino effect of recursion, it can be 
abstracted to follow these steps:
\begin{enumerate}
  \item Read Stanford CoreNLP parse into memory using Phrase classes.  
  \item Identify coded actors in noun phrases.
  \item Identify the usage of the verbs in the verb phrases based on the 
  dictionary entries.
  \item Identify how verbs interact with their constituent verb, prepositional, and noun 
  phrases.
  \item Identify how verbs interact with the noun phrases in their subject 
  position.
  \item Resolve verb+verb interactions.
  \item Return the coding of the uppermost VerbPhrase using CAMEO\footnote{\url{http://eventdata.parusanalytics.com/cameo.dir/CAMEO.09b6.pdf}}\footnote{Schrodt et al. 2008} verb and actor codes, if it satisfies the 
  conditions specified by the user
\end{enumerate}
\section{Classes and class-specific flow}
\subsection{Noun Phrases}
The NounPhrase class only has one unique method, $check\_date()$, which is what 
decides which actor code to choose when the code for a specific person or 
country changes over time. This is taken almost directly from the older version 
of Petrarch. 

The $get\_meaning()$ method in the noun phrase both matches the patterns for the 
actors and agents of word-level children, and combines the meanings of 
constituent PP, NP, and VP children. The priority is given as $Word Patterns > NP > PP 
> VP$, and only when actors and agents are not coded will the node finding the 
meaning look at a lower-priority phrase. This means that the noun phrase 
``American troops in Iraq'' would only code as USAMIL but ``Troops in Iraq'' would code as 
IRQMIL. 
\subsubsection{Pronouns}
When Petrarch encounters a pronoun, it looks up the tree for an antecedent within the same 
sentence. If the pronoun is relfexive (ends with -self or -selves), Petrarch 
looks until it finds a noun, or until it finds a verb phrase with a defined 
subject, and assigns that as the meaning. However, if the pronoun is not 
reflexive,  Petrarch moves up until it finds an S-level phrase, then begins its 
search. This is based on the binding rules that pronouns follow in Generative Grammar. 
Because of the distinction between the two types of pronouns, Petrarch can 
correctly identify that ``itself'' in \textit{A said B hurt itself} refers to B, 
while ``it'' in \textit{A said B hurt it} refers to A. 

Since Petrarch currently has no concept of number or gender, it sometimes makes mistakes
in instances where the pronoun reference depends on the characteristics of the 
nouns in the sentence. Such is the case differentiating \textit{Obama told Hillary 
that he should run for President again} from \textit{Obama told Hillary that she should run
for President again}. Both of these would be interpreted by Petrarch as  \textit{Obama told Hillary that Hillary should run
for President again}

\subsection{Prepositional Phrases}
The $get\_meaning()$ method of PrepPhrase objects returns the meaning of their 
non-preposition constituent. This makes it easier for the actor searching to 
pass through prepositional phrases. The preposition is stored as an attribute of 
the object and is used in some cases to determine whether or not a certain PrepPhrase should 
be considered. 
\subsection{Verb Phrases}
Verb phrases drive most of the complex logic of the program. They play the 
largest role in all three parts of finding ``who did what to whom'', assigning verb 
codes and finding the appropriate noun phrases to fit. The $get\_meaning()$ 
method of verb phrases relies on three other verb-specific methods:
\subsubsection{$get\_upper()$}
This method is fairly simple. If the VP has an NP specifier \footnote{In Syntax, two phrase-level 
siblings are called specifiers. These occur most frequently between VP's,NP's,and PP's. The NP specifier 
of a VP is the phrase that contains the grammatical subject of the verb.} with a coded 
actor, it returns this. Otherwise, this returns nothing. 
\subsubsection{$get\_lower()$}
This is slightly more complicated. In most cases, the verb $get\_lower()$ method 
behaves very similarly to the NounPhrase $get\_meaning()$ method. It looks for 
some coded actor in noun or prepositional phrases, and returns this. 

However, if 
a VP has a VP as a child, it returns the meaning of only that phrase, as well as looking 
for some sort of negation word. The only 
VP's with VP children are modals (could, might, will, etc.) or helping verbs (has, is, 
do, etc.)\footnote{If it intuitively seems like a verb would have another verb phrase as a child, but it does not fall into one of these categories,
it most likely takes a sentence as a child, rather than a verb phrase. }. These won't have other NP, or PP children that are relevant to 
this verb, but can have ``not'' as a child, so this is where negation is 
flagged.
\subsubsection{$get\_code()$}
This is where the program looks to see if the verb follows a pattern 
specified in the dictionary. The patterns consist of four parts: \begin{enumerate}
\item Pre-verb noun phrases \item Pre-verb prepositional phrases \footnote{I can't think of a scenario where 
this would actually be necessary, but the option is there for consistency's sake.} \item 
Post-verb noun phrases \item Post-verb prepositional phrases \end{enumerate}

The process from this point differs for active and passive verbs, but only in 
where each search takes place. Active verbs look for (1) at the closest S-level (Sentence node) above the verb, i.e. the 
nearest point where there will be an NP specifier. It first finds this level via 
the $get\_S()$ method, and looks to see if the head of the NP specifier is part 
of a pattern. If a head is found and there is more to the pattern's noun phrase,
 then the program begins to look for the
rest of the pattern phrase in the noun phrase from which the head came. 
 The verbs find (2) in the same place, but in PP's instead of NP's. Since we 
almost never see patterns with this format in English, this hasn't been fine 
tuned. Then the search begins for (3). This involves checking if any of the 
heads of child NP's are part of a pattern. Then Petrarch follows the same process of looking for 
longer noun patterns within the phrases of the respective heads.
Part (4) looks at child PP's for matches, then matches nouns within the phrases 
if necessary by the same methods it matched child NP's.

For passive verbs, the process for prepositions is exactly the same. However, it 
looks for (1) inside the NP's of child PP's with the preposition ``by,'' ``from,'' or 
``in,''. If no such phrase is found, the verb is left without a subject. This is simply a specific case to deal with how English deals with the 
party that is performing the action in a passive sentence. (3) is found in the 
same place that (1) is found in the active sentences. 

As an illustration, consider the active and passive forms of a simple sentence that would match the pattern
$$ protesters * monument \hspace{4mm} [145]\hspace{4mm} \# DESTROY  $$
 \begin{enumerate} \item The protesters destroyed the 
monument.
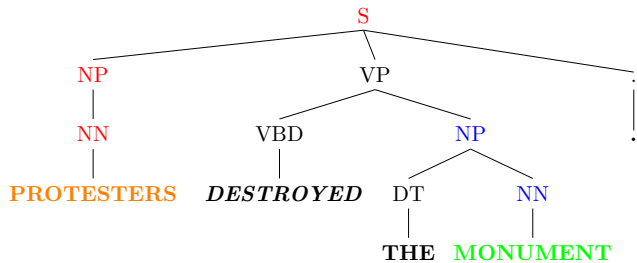
\vspace{10mm}

                    \begin{tikzpicture}[scale= .75]
                    \Tree [.\textcolor{red}{S}  [.\textcolor{red}{NP}    [.\textcolor{red}{NN} {\bf \textcolor{orange}{PROTESTERS}}   ] ] 
[.VP  [.VBD {\bf \emph{ DESTROYED}}  ] [.\textcolor{blue}{NP}  [.DT {\bf THE}  ] [.\textcolor{blue}{NN} {\bf \textcolor{green}{MONUMENT}}  ]   ] 
 ] 
[.. {\bf .}  ]  ] \end{tikzpicture}

   \vspace{10mm}                 
           
\item The monument was destroyed by protesters. 

                   
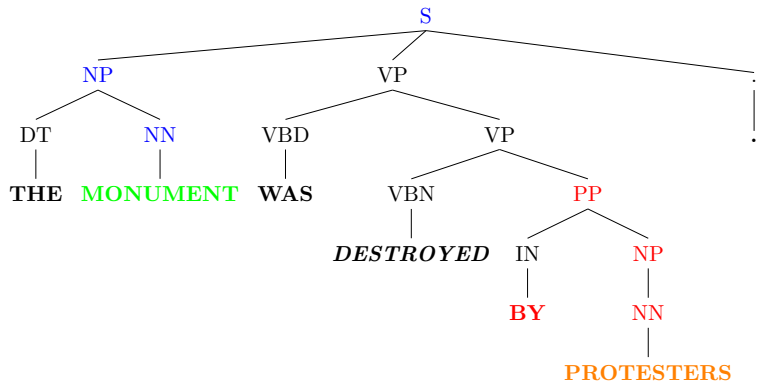
\begin{tikzpicture}[scale= .75]
                    \Tree [.\textcolor{blue}{S}  [.\textcolor{blue}{NP}  [.DT {\bf THE}  ] [.\textcolor{blue}{NN} {\bf \textcolor{green}{MONUMENT}}  ]  ] 
[.VP  [.VBD {\bf WAS}  ] [.VP  [.VBN {\bf \emph{DESTROYED} } ] [.\textcolor{red}{PP}  [.IN {\bf \textcolor{red}{BY}}  ] 
[.\textcolor{red}{NP}    [.\textcolor{red}{NN} {\bf \textcolor{orange}{PROTESTERS}}  ]  ]  
 ]  ] 
 ] 
[.. {\bf .}  ]  ] \end{tikzpicture}
                   
           \vspace{10mm}         

\end{enumerate}

Key: \textcolor{red}{(1) Location} \textcolor{orange}{(1) Match} 
\textcolor{blue}{(3) Location} \textcolor{green}{(3) Match}

Note that this is only for matching patterns entered in the dictionary, not Source and Target matching. 
That happens within the $get\_meaning()$ method, based on the outcomes of $get\_upper()$ 
and $get\_lower()$.
\subsubsection{$get\_meaning()$}
The $get\_meaning()$ method of the Verb class first combines the values of the 
previous three methods in one of a number of ways, depending on what those 
methods find. In most cases, this method returns a list of events that are described by the 
subtree of which the verb phrase is the root. Sometimes, however, if there isn't 
enough verb information available, it will simply return the list of actors described by 
the subtree. In deciding what to do, the verb has several things to consider: 
\begin{itemize}
\item Do I have a source actor? (from $get\_upper()$)
\item Do I have a code? (from $get\_code()$)
\item Do I have a S-level or VP child? (from $get\_lower()$)
\item If so, does that child code an event?
\item If so, how does the event that I code relate the event that it codes?
\end{itemize}

\subsubsection{$match\_transform()$}
This method accounts for the fact that ontologies don't always 
line up exactly with how words work. For example, there are times when you get 
a sentence like ``A says it will attack B,'' but what you're looking to code is ``A threatens B.''
 $Match\_transform()$ reads transformations from the Verb dictionary and checks 
 to see if any of the events match the transformation format. If that's the 
 case, then the event is converted into a simple (S,T,V) format. The entry in 
 the dictionary for that example would be $$ a  \hspace{3mm}(a \hspace{3mm} b \hspace{3mm} WILL\_ATTACK)\hspace{3mm} SAY = a\hspace{3mm} b\hspace{3mm} 138 $$ 
 which is basically post-fix notation. This is described in more detail in the dictionary 
 specifications.
 
\subsubsection{$is\_valid()$}
This method is used to catch a consistent mistake that happens in CoreNLP when a 
past participle is used as an adjective in front of a noun, but is instead coded 
as a verb. 

\subsection{Event extraction}
One call to the $get\_meaning()$ method of the uppermost VP will cause the rest 
of the tree to be parsed, and return the event coding of that VP, which is the 
event coding of the whole tree. Since not all events of the sentence at this 
point might not be complete, the Sentence object which contains the Phrase tree
will call $get\_meaning()$ in its $get\_events()$ method, and check to see if 
the event is satisfactory. If the event that is
returned by $get\_meaning()$ is a complete coding (has all three parts), it is assigned to the 
sentence and the process is complete.

\section{Dictionaries}
Petrarch uses the same Actor, Agent, Discard, and Issue dictionaries as it 
always has, but the newest version has brought changes to the format and 
structure of the Verb Dictionary. The sets of synonymous nouns (synsets) remain 
the same, as well as how the base verbs are organized and stored. The two 
biggest differences are the transformations, which $match\_transform()$ looks at,
and the patterns for matching phrases. 

\subsection{Patterns}
The patterns in the dictionaries should now follow a few simple rules:
\begin{enumerate}
  \item The intended pattern should contain exactly one verb: the verb being 
  matched
  \item The pattern entries should be minimal, i.e. the smallest 
  amount of information necessary to capture the intended phrases. This is just 
  to keep the dictionary small but effective.
  \item The pattern has up to four parts: Pre-verb nouns, Pre-verb Prepositions,
  Post-verb nouns, Post-verb prepositions. 
\end{enumerate}
The patterns themselves also contain additional annotative symbols to provide 
the parser with more syntactic information:
\begin{itemize}
  \item Unmarked words are nouns or particles. These nouns are phrase heads.
  \item \{Bracketed phrases\} are for specifying things that can't be covered by 
  a single noun, e.g. (necessary) adjectives, complex nouns, etc.. The last word in the brackets should be the head. 
  \item Prepositional phrases are (in parentheses). The first word is the 
  preposition, the rest is considered as nouns are.
  \item Note that these prepositional markers can be combined (with \{Noun Phrases\})
\end{itemize}
\subsection{Verb+Verb interaction}
\subsubsection{Combinations}
Verbs can interact with each other in one of two ways. The first is what we 
call a combination. This is what happens when the meaning of the two verbs 
together is literally the meanings of the two verbs individually. These occur 
mostly frequently to specify the subcode of somewhat vague or high-level 
CAMEO categories, like \textit{appeal, intend, refuse} or \textit{demand}.
This is handled using an internal translation of CAMEO codes into a system that
expands the hierarchy of CAMEO beyond the basic top-level/subcode classification
system. This allows for more controllable processing of verb combinations that 
are inherent in CAMEO. So rather than a system where``Intend [030] + Help [070] =Intend to help[033],''
we get ``Intend [3000] + Help [0040] =Intend to help[3040].'' The full 
conversion schema can be found in the \textit{utilities.py} file under 
$convert\_code()$. 

Codes are converted and stored as four-digit hex (base 16) codes.
The general principle behind it is in the table below. The 
first three columns encompass the top-level codes, the fourth position is a specifier. For the 
most part they follow the descriptions here, but some top-level codes have unique subclasses,
which don't follow these specifically. Notice that not all 
combinations refer to CAMEO codes. This is intentional, and means that if we 
wanted to code things beyond CAMEO we could. The strength of this is 
predictability and the possibility of semantic addition. When returning the 
event code, Petrarch converts back to CAMEO for the sake of 
reverse compatibility.

\vspace{12mm}
\begin{tabular}{llll}
             0        &        0   &       0               &        0 \\
            1 Say  &      1 Reduce &   1 Meet   &               1 Leadership \\
            2 Appeal     &    2 Yield  &  2 Settle      &          2 Policy\\
            3 Intend     &~   &            3 Mediate    &           3 Rights\\
            4 Demand      &~     &        4 Aid               &    4 Regime\\
            5 Protest	  &~    &          5 Expel         &        5 Econ\\
            6 Threaten              &~     &    6 Pol. Change    &       6 Military\\
            7 Disapprove    &~    &        7 Mat. Coop  &           7 Humanitarian\\
            8 Posture       &~        &    8 Dip. Coop     &        8 Judicial\\
            9 Coerce          &~      &    9 Assault	        &    9 Peacekeeping\\
            A Investigate &~         &     A Fight		  &  	A Intelligence\\
            B Consult  	&~	&   B Mass violence	&	B Admin. Sanctions\\
           ~ 	&~	&				  ~  			       &     C Dissent\\
           ~ 	&~	&				 ~   				  &      D Release\\
            ~	&~&					~	     		       &     E Int'l Involvement\\
           ~ &~&						  ~ 		            & F De-escalation\\

\end{tabular}

\vspace{12mm}
The one class not present here is 120, which classifies rejections and refusal 
to cooperate. Because the action of ``refusing to do X'' is so often the same as 
``not doing X,'' these are simply categorized as the value of their 
cooperative version minus 0xFFFF. So, since ``provide aid'' is 0040, ``refuse to provide 
aid'' is 0040-FFFF = -FFBF. This is functionally equivalent to the negative, 
since there is no positive FFFF code, the subtraction always yields a negative value. This allows us 
to convert negations such as ``WILL NOT HELP''  = $0 - FFFF + 0040 = -FFBF $ = ``REFUSE TO HELP.'' 
\subsubsection{Transformations}
Sometimes this is insufficient, like when the meaning of the verb interaction 
depends also on the relationships between the nouns that are acting and being 
acted upon. The difference between ``A says B attacked C'' and ``A says A 
attacked B'' is such a case. The first is equivalent to ``A blames B for an 
attack,'' and the second ``A takes credit for an attack on B.'' Since this 
depends on the nouns involved, we must consider them in the transformation
category and not the combination category. The specification on how these 
are formatted is in the documentation.

\section{Example}
Consider the sentence 
\begin{itemize} \item``Israel said a mortar bomb was launched at it from the 
Gaza strip on Tuesday''\end{itemize}
Petrarch would code this sentence as \textit{ISR PSEGZA 112} with the following tree:
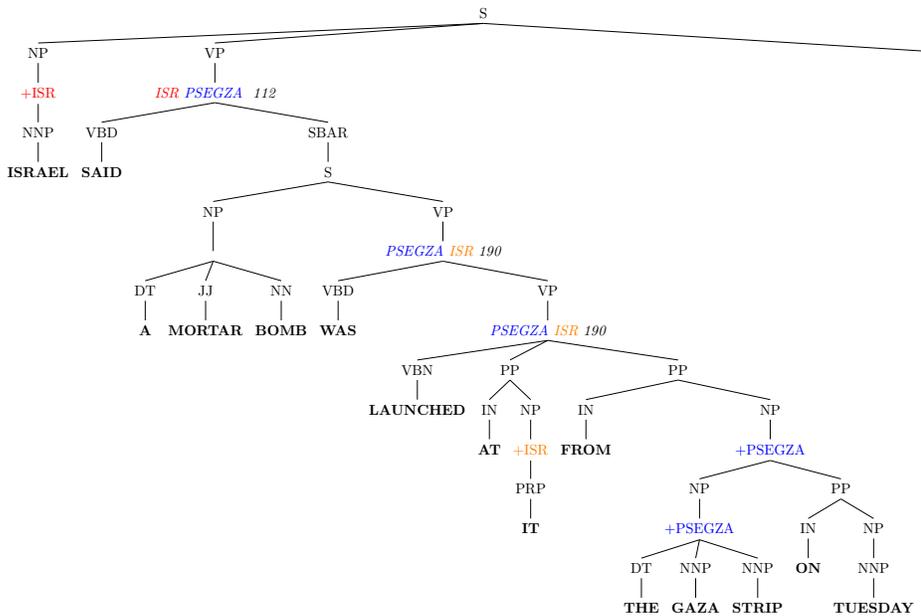
\footnote{For those unfamiliar with CAMEO verb codes, 190 is an organized attack, while 112 is an accusation of aggression}\\
                    \begin{tikzpicture}[scale= .5]
                      
                    \Tree [.S  [.NP  [.\color{red}{+ISR} [.NNP {\bf ISRAEL}  ]  ] ] 
[.VP  [.{\it \color{red}{ISR} \color{blue}{PSEGZA} \color{black}{ 112}} [.VBD {\bf SAID}  ] [.SBAR  [.S  [.NP  [.{} [.DT {\bf A}  ] [.JJ {\bf MORTAR}  ] [.NN {\bf BOMB}  ]  ] ] 
[.VP  [.{\it \color{blue}{PSEGZA} \color{orange}{ISR} \color{black}{190}} [.VBD {\bf WAS}  ] [.VP  [.{\it \color{blue}{PSEGZA} \color{orange}{ISR} \color{black}{190}} [.VBN {\bf LAUNCHED}  ] [.PP  [.IN {\bf AT}  ] [.NP  [.\color{orange}{+ISR} [.PRP {\bf IT}  ]  ] ] 
 ] [.PP  [.IN {\bf FROM}  ] [.NP  [.\color{blue}{+PSEGZA} [.NP  [.{\color{blue}{+PSEGZA}} [.DT {\bf THE}  ] [.NNP {\bf GAZA}  ] [.NNP {\bf STRIP}  ]  ] ] 
[.PP  [.IN {\bf ON}  ] [.NP  [.NNP {\bf TUESDAY}  ]  ] ] 
  ] ] 
 ]  ] ] 
 ]  ] ] 
 ] ] 
   ]  
[.. {\bf .}  ]  ]   \end{tikzpicture}

The color coding shows where the actor codes come from. The significant steps taken in this parse involve the verbs ``said'' and 
``flaunched,'' and the pronoun ``it.'' The pronoun coreference follows the 
non-reflexive matching process described above. When Petrarch is analyzing 
``launched,'' it \begin{enumerate} 
\item Identifies the verb as passive 
\item Finds the patterns for this verb
\item Finds the target under the prepositional phrase with ``it'' 
\item Identifies the antecedent of  ``it'' to be ``ISR''  
\item Finds the source under the prepositional phrase with ``from''
\end{enumerate}
Then, the analysis of ``said'' follows the process:
\begin{enumerate}
  \item Finds the lower event (PSEGZA ISR 190)
  \item Identifies the subject of ``said'' as ISR
  \item Matches this with the dictionary-specified verb transformation\\ \textit{a (b . ATTACK) SAY} 
  = a b 112
  \item Transforms this into \textit{ISR PSEGZA 112}.
\end{enumerate}

\section{Works Cited}

\end{document}